\documentclass[a4]{article}

\usepackage{hyperref}
\usepackage{subcaption}
\usepackage{amsmath}
\usepackage{amsfonts}
\usepackage{authblk}
\usepackage{graphicx}
\usepackage{booktabs}
\usepackage{multirow}
\begin{document}
	
	\title{\textbf{A Comparison of the Delta Method and the Bootstrap in Deep Learning Classification}}
	\author[1,2]{Geir K. Nilsen}
	\author[1]{Antonella Z. Munthe-Kaas}
	\author[1]{Hans J. Skaug}
	\author[1]{Morten Brun}
	\affil[1]{Department of Mathematics, University of Bergen}
	\affil[2]{geir.kjetil.nilsen@gmail.com}
	\date{}                     %% if you don't need date to appear
	\setcounter{Maxaffil}{0}
	\renewcommand\Affilfont{\itshape\small}

	%\address{Department of Mathematics, University of Bergen}
	
	
	
	
	\maketitle

\begin{abstract}
We validate the deep learning classification adapted Delta method introduced in \cite{nilsen2} by a comparison with the classical Bootstrap. We show that there is a strong linear relationship between the quantified predictive epistemic uncertainty levels obtained from the two methods when applied on two LeNet-based neural network classifiers using the MNIST and CIFAR-10 datasets. Furthermore, we demonstrate that the Delta method offers a five times computation time reduction compared to the Bootstrap.
\end{abstract}

\section{Introduction}
It can be beneficial to distinguish between epistemic and aleatoric uncertainty in machine learning models \cite{hullermeier}. Bayesian statistics provides a coherent framework for representing epistemic uncertainty in neural networks \cite{mackay}, but has not so far gained widespread use in deep learning \cite{goodfellowetal} -- presumably due to the high computational cost that traditionally comes with Fisher information based methods. In particular, the Delta method \cite{hoef, khosravi} depends on the empirical Fisher information matrix which grows quadratically with the number of neural network parameters $P$ -- and its direct application in modern deep learning is therefore prohibitively expensive. To mitigate this, \cite{nilsen2} proposed a low cost variant of the Delta method applicable to $L_2$-regularized deep neural networks based on the top $K$ eigenpairs of the Fisher information matrix.

In this paper, we validate the methodology introduced in \cite{nilsen2} by a comparison with the classical Bootstrap \cite{efron, khosravi, laksh, osband, osband2}. We show that there is a strong linear relationship between the quantified epistemic uncertainty levels obtained from the two methods when applied on two LeNet-based neural network classifiers using the MNIST and CIFAR-10 datasets.

The paper is organized as follows: in Section \ref{sec:intro} we review the Bootstrap and the Delta method in a deep learning classification context. In Section \ref{sec:classifiersdef} we introduce two LeNet-based classifiers which will be used in the comparison in Section \ref{sec:comp}, and finally, in Section \ref{sec:summary} we summarize the paper and give some concluding remarks.

\section{Introduction to the Methodologies}\label{sec:intro}
In the following, we denote the training set by $\lbrace x_n\in\mathbb{R}^{T_1},y_n\in\mathbb{R}^{T_L}\rbrace_{n=1}^{N}$, the test set by $\lbrace x_n\in\mathbb{R}^{T_1},y_n\in\mathbb{R}^{T_L}\rbrace_{n=1}^{N_{\text{test}}}$ and an arbitrary input example by $x_0$. The parameter space is denoted by the vector $\omega\in\mathbb{R}^{P}$, where $P$ is the number of parameters (weights and biases) in the model. The parameter values after training is denoted by the vector $\hat{\omega}\in\mathbb{R}^P$. Furthermore, a prediction for $x_0$ is denoted by $\hat{y}_0 = f(x_0, \hat{\omega})\in\mathbb{R}^{T_L}$ where $f: \mathbb{R}^{T_1\times P} \rightarrow \mathbb{R}^{T_L}$ is a deep neural network model function \cite{goodfellowetal} and where $T_L$ denotes the number of classes. Furthermore, it is assumed that the cost function denoted by $C$ is $L_2$-regularized with a regularization-rate factor $\lambda/2$.

\subsection{The Bootstrap in Deep Learning Classification}
In the context of deep learning classification, the classical Bootstrap method starts by creating $B$ datasets from the original dataset by sampling with replacement. Subsequently, $B$ networks are trained separately on each of the bootstrapped datasets. The epistemic uncertainty for each of the $T_L$ class predictions (in standard deviations) associated with prediction of $x_0$ is obtained by the sample standard deviation over the ensemble of $B$ predictions,
\begin{equation}
	\widetilde{\sigma}_{\text{boot}}(x_0) = \sqrt{\frac{1}{B-1}\sum_{b=1}^{B}(\hat{y}_{0}^{(b)} - \overline{\hat{y}_{0}})^2} \in\mathbb{R}^{T_L},
	\label{eq:bootstrapsigma}
\end{equation}
where the vector $\hat{y}_{0}^{(b)}$ represents the $T_L$ predictions for $x_0$ (one probability per class) obtained from the $b$th bootstrapped network, and where $\overline{\hat{y}_{0}}$ is the sample mean, 
\begin{equation}
	\overline{\hat{y}_{0}} = \frac{1}{B}\sum_{b=1}^{B}\hat{y}_{0}^{(b)} \in\mathbb{R}^{T_L}.
\end{equation}
The method is easy to implement efficiently in practice. Training $B$ networks is an `embarrassingly' parallel problem, and the space complexity for the bootstrapped datasets is just $O(BN)$ when an indexing scheme is used for the sampling with replacement. The experiments conducted in this paper is based on the example \texttt{pydeepboot.py} provided in the  \texttt{pydeepdelta} provision \cite{pydeepdeltamodule}.

\subsection{The Delta Method in Deep Learning Classification}
The Delta method was adapted to the deep learning classification context by \cite{nilsen2}. The adaption addresses several fundamental difficulties that arise when the method is applied in deep learning. In essence, it is shown that an approximation of the eigendecomposition of the Fisher information matrix utilizing only $K$ eigenpairs allows for an efficient implementation with bounded worst-case approximation errors. We briefly review the standard method here for convenience.

An approximation of the epistemic component of the uncertainty associated with the prediction of $x_0$ can be found by the formula
\begin{equation}
	\widetilde{\sigma}_{\text{delta}}(x_0) = \sqrt{\text{diag}\big(F\Sigma F^T\big)}\in \mathbb{R}^{T_L}\label{eq:sigma1},
\end{equation}
where the sensitivity matrix $F$ in (\ref{eq:sigma1}) is defined
\begin{equation}
	F = \begin{bmatrix}F_{ij}\end{bmatrix}\in \mathbb{R}^{T_L\times P},~F_{ij} = \frac{\partial}{\partial \omega_j} f_i(x_0, \omega)\bigg\rvert_{\omega=\hat{\omega}}.\label{eq:F}
\end{equation}
The covariance matrix $\Sigma$ in \eqref{eq:sigma1} can be estimated by several alternative estimators. In \cite{nilsen2} it was demonstrated that the Hessian estimator, the Outer-Products of Gradients (OPG) estimator and the Sandwich estimator lead to nearly perfect correlated results for two different deep learning models. Since the models discussed in this paper are identical to those in \cite{nilsen2}, we thus focus only on one of the estimators, namely the OPG estimator defined by
\begin{equation}
	\Sigma = \frac{1}{N}G^{-1} = \frac{1}{N}\left[\frac{1}{N}\sum_{n=1}^N \frac{\partial C_n}{\partial \omega} \frac{\partial C_n}{\partial \omega}^T\bigg\rvert_{\omega=\hat{\omega}} + \lambda I\right]^{-1}\in \mathbb{R}^{P\times P},\label{eq:opg}
\end{equation}
where the summation part of $G$ corresponds to the empirical covariance of the gradients of the cost function evaluated at $\hat{\omega}$. As discussed in \cite{nilsen2}, the term $\lambda I$ is explicitly added in order to make the OPG estimator asymptotically equal to the Hessian estimator, as is the primary motivation for the former as a plug-in replacement of the latter in the first place.

When the Delta method is implemented under the framework of \cite{nilsen2}, it has several desirable properties: a) requires only $O(PK)$ space and $O(KPN)$ time, b) fits well with deep learning software frameworks based on automatic differentiation, c) works with any $L_2$-regularized neural network architecture, and d) does not interfere with the training process as long as the norm of the gradient of the cost function is approximately equal to zero after training.

\section{The Neural Network Classifiers}\label{sec:classifiersdef}
We deploy two LeNet-based neural network architectures which differs only by the number of neurons in two of the layers in order to individually match the formats of the MNIST and CIFAR-10 datasets. Our TensorFlow code for the Delta method is based on the \texttt{pydeepdelta} Python module \cite{pydeepdeltamodule}, and is fully deterministic \cite{nagarajan}. The corresponding Bootstrap implementation can be found in the same repository.

\subsection{MNIST}
There are $L=6$ layers, layer $l=1$ is the input layer represented by the input vector. Layer $l=2$ is a $3\times 3\times 1\times 32$ convolutional layer followed by max pooling with stride equal to $2$ and with a ReLU activation function. Layer $l=3$ is a $3\times 3\times 32\times 64$ convolutional layer followed by max pooling with a stride equal to $2$, and with ReLU activation function. Layer $l=4$ is a $3\times 3\times 64\times 64$ convolutional layer with ReLU activation function. Layer $l=5$ is a $576\times 64$ dense layer with ReLU activation function, and the output layer $l=6$ is a $64\times T_L$ dense layer with softmax activation function, where the number of classes (outputs) is $T_L=10$. The total number of parameters is $P=93322$.

\subsection{CIFAR-10}
There are $L=6$ layers, layer $l=1$ is the input layer represented by the input vector. Layer $l=2$ is a $3\times 3\times 3\times 32$ convolutional layer followed by max pooling with stride equal to $2$ and with a ReLU activation function. Layer $l=3$ is a $3\times 3\times 32\times 64$ convolutional layer followed by max pooling with a stride equal to $2$, and with ReLU activation function. Layer $l=4$ is a $3\times 3\times 64\times 64$ convolutional layer with ReLU activation function. Layer $l=5$ is a $1024\times 64$ dense layer with ReLU activation function, and the output layer $l=6$ is a $64\times 10$ dense layer with softmax activation function, where the number of classes (outputs) is $T_L=10$. The total number of parameters is $P=122570$.

\subsection{Training Details}
For the Bootstrap networks, we test two different weight initialization variants: dynamic random normal weight initialization (DRWI) and static random normal weight initialization (SRWI). The former uses a different (e.g. dynamic) seed across the replicates, meaning that each network in the DRWI Bootstrap ensemble will start out with different random weight values. The latter case uses the same (e.g. static) seed across the replicates, and hence all the networks in the SRWI Bootstrap ensemble receives the same random initial weight values. For all networks, we use zero bias initialization. Futhermore, to investigate the impact of random weight initialization on the Delta method, we apply the Delta method 16 times on a set of 16 networks distinguished only by DRWI.

We use the cross-entropy cost function with a $L_2$-regularization rate $\lambda=0.01$, and utilize the Adam \cite{kingma, bottou} optimizer with a batch size of $100$, and no form of randomized data shuffling. To ensure convergence (e.g. $||\nabla C(\hat{\omega})||_2 \approx 0$), we apply two slightly different learning rate schedules given by the following (step, rate) pairs: MNIST = $\lbrace(0, 10^{-3}), (60\text{k}, 10^{-4}), (70\text{k}, 10^{-5}), (80\text{k}, 10^{-6})\rbrace$ and CIFAR-10 = $\lbrace(0, 10^{-3}), (55\text{k}, 10^{-4}), (85\text{k}, 10^{-5}), (95\text{k}, 10^{-6}, (105\text{k}, 10^{-7})\rbrace$. For MNIST, we stop the trainings after $90,000$ steps, while for CIFAR-10, after $115,000$ steps -- corresponding to the overall training statistics shown in Table \ref{tab:trainingstats}.
\begin{table}[hp]
	\scalebox{0.85}{
		\begin{tabular}{c|l|l|l|l|l}
			\textbf{Networks} & \textbf{Dataset} & \textbf{\shortstack{\\Training Set \\ Accuracy}} & \textbf{\shortstack{\\Test Set \\ Accuracy}} & \textbf{$C(\hat{\omega})$} & \textbf{$||\nabla C(\hat{\omega})||_2$} \\ \hline
			\multirow{2}{*}{\shortstack{\\DRWI\\Bootstrap\\B=100}} & MNIST            & $0.979 \pm 0.000$              & $0.981 \pm 0.001$          & $0.253 \pm 0.006$              & $0.016 \pm 0.013$                            \\ \cline{2-2} 
			& CIFAR-10         & $0.705 \pm 0.025$              & $0.684 \pm 0.020$          & $1.248 \pm 0.042$              & $0.035 \pm 0.020$                                                                                                                              \\ \cline{1-2}
			\multirow{2}{*}{\shortstack{\\SRWI\\Bootstrap\\B=100}} & MNIST            & $0.979 \pm 0.000$              & $0.981 \pm 0.001$          & $0.254 \pm 0.002$              & $0.017 \pm 0.013$                                                                       \\ \cline{2-2} 
			& CIFAR-10         & $0.715 \pm 0.010$              & $0.693 \pm 0.009$          & $1.235 \pm 0.018$              & $0.031 \pm 0.014$                                                                                                                              \\ \cline{1-2}
			\multirow{2}{*}{\shortstack{\\Delta\\16 reps\\(DRWI)}}     & MNIST            & $0.979 \pm 0.000$              & $0.981 \pm 0.001$        & $0.257 \pm 0.002$              & $0.016 \pm 0.005$                                                   \\ \cline{2-2} 
			& CIFAR-10         & $0.701 \pm 0.032$              & $0.687 \pm 0.029$        & $1.284 \pm 0.053$            & $0.030 \pm 0.012$                                                                                                        \\ %\cline{1-1}
		\end{tabular}
	}
	\caption{Training statistics for the Delta and Bootstrap networks. The DRWI and SRWI Bootstrap ensembles each consists of $B=100$ bootstrapped networks, while the Delta method is applied repeatedly on 16 networks distinguished only by DRWI. Averages $\pm$ two standard deviations are calculated across the $B=100$ networks for the Bootstrap, and across the 16 repetitions for the Delta method.}
	\label{tab:trainingstats}
\end{table}

\section{Comparison}\label{sec:comp}
\begin{figure}[htp]
	\centering
	\includegraphics[scale=0.33]{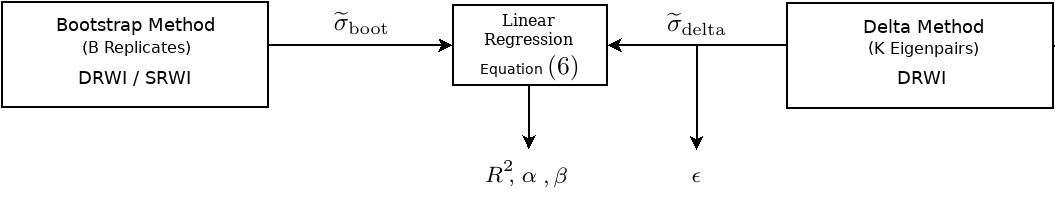}
	\caption{Regression \eqref{eq:regression} of $\widetilde{\sigma}_\text{boot}$ onto $\widetilde{\sigma}_\text{delta}$.}
	\label{fig:testdesign}
\end{figure}

The basic comparison design entails a set of 16 linear regressions on the predictive uncertainty estimates obtained from the two methods using test sets as input data
\begin{align}
	\widetilde{\sigma}_{\text{boot}}(x_n)_{m} = \alpha_{d} + \beta_{d} \widetilde{\sigma}_{\text{delta}}(x_n)_{m,d} + e_{n,m,d}, 
	\quad n &= 1,2,\hdots,N_{\text{test}}\nonumber\\
	\quad m &= 1,2,\hdots,T_L\nonumber\\
	\quad d &= 1,2,\hdots,16.\label{eq:regression}
\end{align}
Accounting for the two variants of the Bootstrap (SRWI/DRWI), this leads to two sets of squared correlation coefficients, intercepts, slopes and Delta method approximation errors, respectively denoted by $\lbrace R^{2}_{d}, \alpha_{d}, \beta_{d}, \epsilon_{d}\rbrace_{d=1}^{16}$. Furthermore, as we wish to analyze the impact of the number of Bootstrap replicates and the number of Delta method eigenpairs, we generate these sets for various $B$ and $K$. An outline of the setup is shown in Figure \ref{fig:testdesign}. 

\begin{figure}[htp]
	\centering
	\begin{subfigure}{.5\textwidth}
		\includegraphics[scale=0.43]{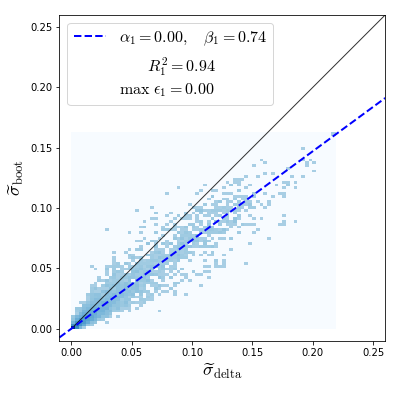}
		\caption{MNIST}
		\label{fig:mnisttestuncertaintyBregression}
	\end{subfigure}%
	\begin{subfigure}{.5\textwidth}
		\includegraphics[scale=0.43]{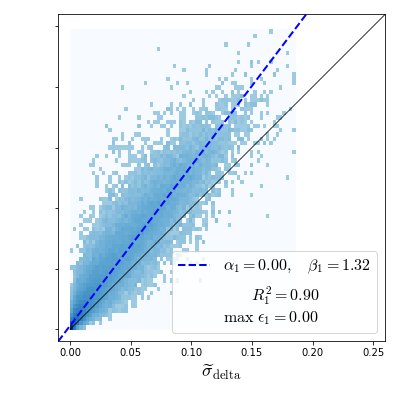}
		\caption{CIFAR-10}
		\label{fig:cifar-10testuncertaintyBregression}
	\end{subfigure}
	\caption{Predictive uncertainty estimates obtained from the Delta method (first repetition, $d=1$) against the DRWI Bootstrap for (a) MNIST using $B=100$ replicates and $K=1500$ eigenpairs, and (b) CIFAR-10 using $B=100$ replicates and $K=2500$ eigenpairs.}
	\label{fig:testuncertaintyBregression}
\end{figure}

Figure \ref{fig:testuncertaintyBregression} shows scatter plots of the regression results for the first repetition ($d=1$) of the Delta method against the DRWI Bootstrap ensemble. These plots are based on $B=100$ bootstrap replicates, and we have selected $K=1500$ eigenpairs for MNIST and $K=2500$ eigenpairs for CIFAR-10. Clearly, there is a strong linear relationship between the two methods: the squared correlation coefficients are $R_1^2=0.94$ for MNIST and $R_1^2=0.90$ for CIFAR-10. On the other hand, the absolute uncertainty level differs between the methods and datasets. This can be seen by the slope coefficients, where the Delta method is overestimating ($\beta_1 < 1$) on MNIST, and underestimating ($\beta_1 > 1$) on CIFAR-10. Further, since the estimated intercepts ($\alpha_1$) are zero, there are no offsets between the methods. Finally, we see that the maximum across examples and class outputs of the Delta method approximation errors ($\epsilon_1$) are zero, so there is nothing to be achieved by increasing $K$. As we will see later, $K$ has here been selected unnecessarily high and can be significantly reduced with no loss of accuracy.

\subsection{Discussion of the Regression Results as a Function of $B$ and $K$}
\begin{figure}[htp]
	\begin{subfigure}{0.5\textwidth}
		\centering
		\includegraphics[scale=0.3]{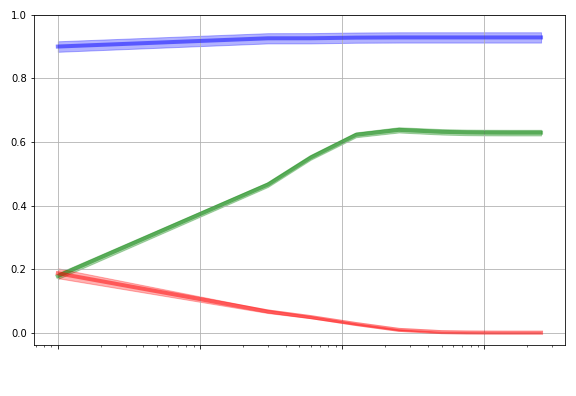}
		\caption{MNIST\\Delta vs. SRWI Bootstrap}
		\label{fig:mnisttestuncertaintyA}
	\end{subfigure}%
	\begin{subfigure}{.5\textwidth}
		\centering
		\includegraphics[scale=0.3]{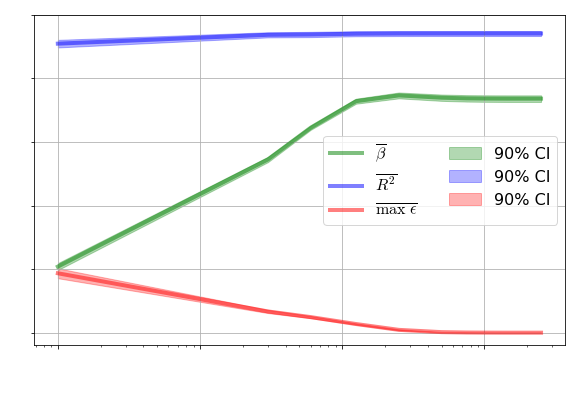}
		\caption{MNIST\\Delta vs. DRWI Bootstrap}
		\label{fig:mnisttestuncertaintyB}
	\end{subfigure}
	\begin{subfigure}{0.5\textwidth}
		\centering
		\includegraphics[scale=0.3]{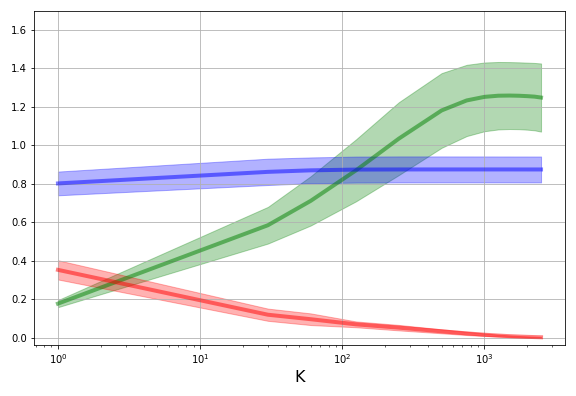}
		\caption{CIFAR-10\\Delta vs. SRWI Bootstrap}
		\label{fig:cifar-10testuncertaintyA}
	\end{subfigure}%
	\begin{subfigure}{0.5\textwidth}
		\centering
		\includegraphics[scale=0.3]{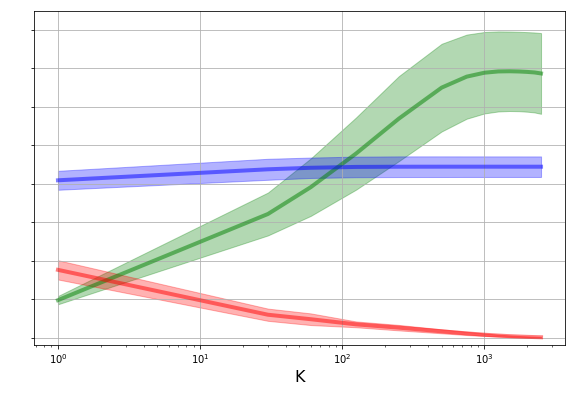}
		\caption{CIFAR-10\\Delta vs. DRWI Bootstrap}
		\label{fig:cifar-10testuncertaintyB}
	\end{subfigure}%
	\caption{Summaries of the regressions of $\widetilde{\sigma}_{\text{boot}}$ onto $\widetilde{\sigma}_{\text{delta}}$ as given by \eqref{eq:regression}, for different values of $K$ and a fixed $B=100$. The solid lines and the associated confidence intervals represent the mean and the variation of the regression results across the 16 repetitions of the Delta method.
}
	\label{fig:uncertainty}
\end{figure}

The results from the full set of regressions ($d=1,2,\hdots,16$) holding a fixed $B=100$ are shown in Figure \ref{fig:uncertainty}. The primary observations are as follows: The mean squared correlation coefficients $R^2$ are generally high for MNIST and CIFAR-10, meaning that there is a strong linear relationship between the uncertainty levels obtained by the Bootstrap and the Delta method. For the lowest $K$, the $R^2$ starts out at 90\% for MNIST, and at 81\% for CIFAR-10. As $K$ grows, an increase by only $4$\% is observed for MNIST, while  8\% for CIFAR-10. The major difference observed as $K$ increases lies in the absolute uncertainty levels expressed by the slope $\beta$: for MNIST, the slope stabilizes at around $K=600$ while at about $K=1000$ for CIFAR-10. The same trend is reflected in the maximum approximation errors $\epsilon$, where we respectively see them approach zero at the same values for $K$. Although not shown in the plots, the regression intercepts $\alpha$ are always zero, meaning that there is no offset in the uncertainty estimates by the two methods. 

The main difference found from applying DRWI opposed to SRWI for the Bootstrap ensembles, is that the absolute level of uncertainty increases with DRWI. This is expected, since the DRWI version of the Bootstrap will be more prone to reaching different local minima, and therefore also captures this additional variance. Supporting evidence for this hypothesis is evident by CIFAR-10's wider confidence intervals. A more pronounced geometry difference across various local minima will ultimately lead to higher variability in the $R^2$ and $\beta$. A slightly higher mean $R^2$ (+1-2\%) is also observed for the DRWI version of the Bootstrap. This is reasonable given the fact that also the Delta method networks are more prone to reaching different local minima across the 16 repetitions because of DRWI.

\begin{figure}[htp]
	\begin{subfigure}{0.5\textwidth}
		\centering
		\includegraphics[scale=0.3]{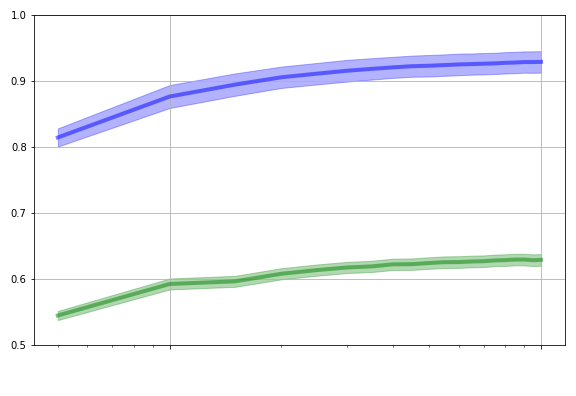}
		\caption{MNIST\\Delta vs. SRWI Bootstrap}
		\label{fig:mnistrainuncertaintyAoverB}
	\end{subfigure}%
	\begin{subfigure}{.5\textwidth}
		\centering
		\includegraphics[scale=0.3]{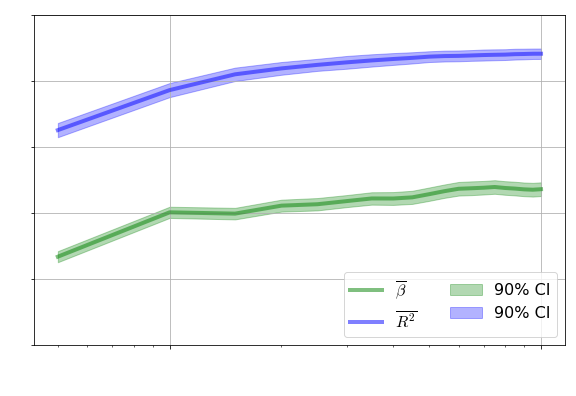}
		\caption{MNIST\\Delta vs. DRWI Bootstrap}
		\label{fig:mnisttrainuncertaintyBoverB}
	\end{subfigure}
	\begin{subfigure}{0.5\textwidth}
		\centering
		\includegraphics[scale=0.3]{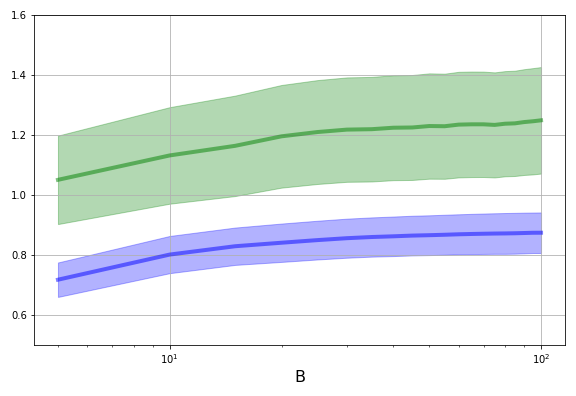}
		\caption{CIFAR-10\\Delta vs. SRWI Bootstrap}
		\label{fig:cifar-10testuncertaintyAoverB}
	\end{subfigure}%
	\begin{subfigure}{0.5\textwidth}
		\centering
		\includegraphics[scale=0.3]{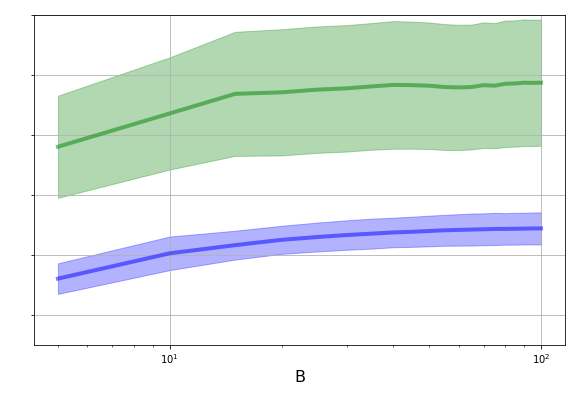}
		\caption{CIFAR-10\\Delta vs. DRWI Bootstrap}
		\label{fig:cifar-10trainuncertaintyBoverB}
	\end{subfigure}%
	\caption{Summaries of the regressions of $\widetilde{\sigma}_{\text{boot}}$ onto $\widetilde{\sigma}_{\text{delta}}$ as given by \eqref{eq:regression}, for different values of $B$ and a fixed number of eigenpairs $K$. The solid lines and the belonging confidence intervals represent the mean and the variation of the regression results across the 16 repetitions of the Delta method.}
	\label{fig:uncertaintyoverB}
\end{figure}

Figure \ref{fig:uncertaintyoverB} shows the same type of comparison when the number of Bootstrap replicates $B$ varies, and the number of eigenpairs are fixed ($K=1500$ for MNIST and $K=2500$ for CIFAR-10). The main observation from this experiment is that there is very little to achieve by selecting a larger ensemble size $B$ than about 50, as this is the point where the mean slope and squared correlation coefficient stabilizes.

\subsection{Computation Time}
Table \ref{tab:timings} shows the computation time for the two methods when executed on a Nvidia RTX 2080 Ti based GPU. For MNIST, the smallest $K$ leading to acceptable approximation errors and stable absolute uncertainty levels for the Delta method is at $K=600$, while for CIFAR-10 the same applies at $K=1000$. Furthermore, the smallest acceptable $B$ leading to stable correlation and absolute uncertainty levels for the Bootstrap is at $B=50$. We conclude that in these experiments the Delta method outperforms the Bootstrap in terms of computation time by a factor $4.6$ on MNIST, and a factor $5.9$ for CIFAR-10.

\begin{table}[htp]
	\scalebox{0.64}{
	\begin{tabular}{c|c|c|c|c|c|c|c}
		\multirow{2}{*}{\textbf{Method}} & \multicolumn{1}{l|}{\multirow{2}{*}{\textbf{Classifier}}} & \multirow{2}{*}{\textbf{B}} & \multirow{2}{*}{\textbf{K}} & \multirow{2}{*}{\textbf{Initial Phase [h:mm:ss]}} & \multicolumn{2}{c|}{\textbf{Prediction Phase [mm:ss]}} & \multirow{2}{*}{\textbf{Total [h:mm:ss]}} \\ \cline{6-7}
		& \multicolumn{1}{l|}{}                                     &                             &                             &                                                   & \textbf{Training Set}       & \textbf{Test Set}       &                                           \\ \hline
		\multirow{2}{*}{Bootstrap}       & MNIST                                                     & \multirow{2}{*}{50}         & \multirow{2}{*}{N/A}        & 4:08:28                                           & 00:19                        & 00:03                    & 4:08:50                                   \\ \cline{2-2}
		& CIFAR-10                                                  &                             &                             & 7:37:16                                           & 00:40                        & 00:07                    & 7:38:04                                   \\ \cline{1-4}
		\multirow{2}{*}{Delta}           & MNIST                                                     & \multirow{2}{*}{N/A}        & 600                         & 0:42:33                                           & 9:52                        & 1:37                    & 0:54:02                                     \\ \cline{2-2}\cline{4-4}
		& CIFAR-10                                                  &                             & 1000                        & 1:00:54                                           & 14:44                       & 02:56                    & 1:18:35                                  
	\end{tabular}
	}
	\caption{Computation time for the Bootstrap and Delta method. For the Bootstrap, the `initial phase' accounts for the parallelized training of $B$ networks, while the `prediction phase' accounts for the predictive epistemic uncertainty estimation \eqref{eq:bootstrapsigma}, which is further divided into the training and test sets. For the Delta method, the `initial phase' accounts for the approximate eigendecomposition of the covariance matrix \eqref{eq:opg}, while the `prediction phase' accounts for the predictive epistemic uncertainty estimation \eqref{eq:sigma1}, further divided into the training set and test sets.}
	\label{tab:timings}
\end{table}

\section{Concluding Remarks}\label{sec:summary}
We have shown that there is a strong linear relationship between the predictive epistemic uncertainty estimates obtained by the Bootstrap and the Delta method when applied on two different deep learning classification models. Firstly, we find that the number of eigenpairs $K$ in the Delta method can be selected order of magnitudes lower than $P$ with no loss of correspondence between the methods. This coincides with the fact that when the Delta method approximation errors are sufficiently close to zero, there is no nothing to achieve by a further increase in $K$, and therefore the correspondence will stabilize at this point.

Secondly, we find that the DRWI version of the Bootstrap yields the best correspondence, and that there is little to achieve by using more than $B=50$ replicates. Thirdly, we observe that the most complex model (CIFAR-10) yields a high variability in the correspondence across multiple DRWI Delta method runs. We interpret this effect as caused by cost functional multi-modality, and that the Delta method fails to capture the additional variance tied to reaching local minima of different geometric characteristics. Finally, in our experiments we have seen that the Delta method outperforms the Bootstrap in terms of computation time by a factor $4.6$ on MNIST and by a factor $5.9$ for CIFAR-10.

\bibliographystyle{abbrv}
\bibliography{references}

\end{document}